\documentclass{article}

\PassOptionsToPackage{numbers,square,compress,comma}{natbib}
 \usepackage[main, final]{neurips_2026}
\usepackage[utf8]{inputenc} % allow utf-8 input
\usepackage[T1]{fontenc}    % use 8-bit T1 fonts
\usepackage{hyperref}       % hyperlinks
\usepackage{url}            % simple URL typesetting
\usepackage{booktabs}       % professional-quality tables
\usepackage{amsfonts}       % blackboard math symbols
\usepackage{nicefrac}       % compact symbols for 1/2, etc.
\usepackage{microtype}      % microtypography
\usepackage{xcolor}         % colors
\usepackage{graphicx}
\usepackage{subcaption}
\usepackage{titlesec}
\usepackage{multirow}
\usepackage[acronym]{glossaries}
\glsdisablehyper
\newacronym{llm}{LLM}{Large Language Model}
\newacronym{ffn}{FFN}{Feed Forward Network}
\newacronym{nlp}{NLP}{Natural Language Processing}

\titlespacing*{\paragraph}
{0pt}   % left margin
{0.5ex}   % space before
{0.5em}   % space after (run-in spacing)

\title{SelectInfer: Selective Neuron Loading and Computation for On-Device LLMs}

\author{
  Huzaifa Shaaban Kabakibo$^{1}$\thanks{Equal contribution.} \\
  \texttt{huzaifa@mail.uni-paderborn.de}
  \And
  Eric Schniedermeyer$^{1}$\footnotemark[1] \\
  \texttt{eric1@mail.uni-paderborn.de}
  \And
  Artem Burchanow$^{1}$\footnotemark[1] \\
  \texttt{artemb@mail.uni-paderborn.de}
  \And
  Lin Wang$^{1}$ \\
  \texttt{lin.wang@uni-paderborn.de}
  \\
  $^{1}$Department of Computer Networks, Paderborn University, Paderborn, Germany
}

\begin{document}

\maketitle

\begin{abstract}
    \glspl{llm} have demonstrated remarkable capabilities across a range of \gls{nlp} tasks, but their high computational and memory demands pose significant challenges for deployment on resource-constrained edge devices. Existing approaches to model compression and optimization often rely on coarse-grained pruning or quantization, which can compromise accuracy or require re-training and fine-tuning. In this work, we introduce \textbf{SelectInfer}, a neuron-level optimization framework that enables efficient \gls{llm} inference on edge devices through selective neuron loading and computation. By profiling and identifying both task-specific and general-purpose neurons using an \textit{offline LLM profiler}, SelectInfer implements two key optimizations: \textit{selective loading}, which reduces memory footprint by selectively loading a subset of neurons that were identified to be most important during the offline stage, and \textit{selective computation}, which dynamically computes only the most relevant neurons at runtime. Evaluation across multiple datasets shows that SelectInfer achieves significant reductions in memory footprint and computation while preserving task performance, making it a practical step towards enabling \gls{llm} deployment on edge devices.
\end{abstract}

% ------------------- Introduction -------------------

\section{Introduction}

Large Language Models (LLMs) have rapidly become foundational tools across a wide array of applications, including natural language writing, code generation, healthcare diagnostics, financial analysis, and many more ~\citep{llama, llm-usecases1, llm-usecases2, llm-usecases3}. Their impressive capabilities have driven their widespread adoption, predominantly through deployment in powerful data centers equipped with high-end GPUs. These centralized environments offer immense computational resources required to run LLM inference efficiently.

However, growing concerns around user privacy, the need for application customization, and the demand for offline capabilities~\citep{chun2011clonecloud, powerinfer2, PrivateGPT_2023, lyu2023llm} have fueled increasing interest in enabling LLM inference directly on edge devices, such as NVIDIA Jetson~\citep{nvidia-jetson-orin-nano} and Google Coral~\citep{googlecoral}.
Edge deployment offers the potential for greater data privacy and responsiveness, but it also faces significant challenges. Unlike data centers, edge devices are constrained by limited memory and computational power, which complicates the direct execution of large, resource-hungry models.

\begin{figure*}[!t]
\centering
\includegraphics[width=0.9\linewidth]{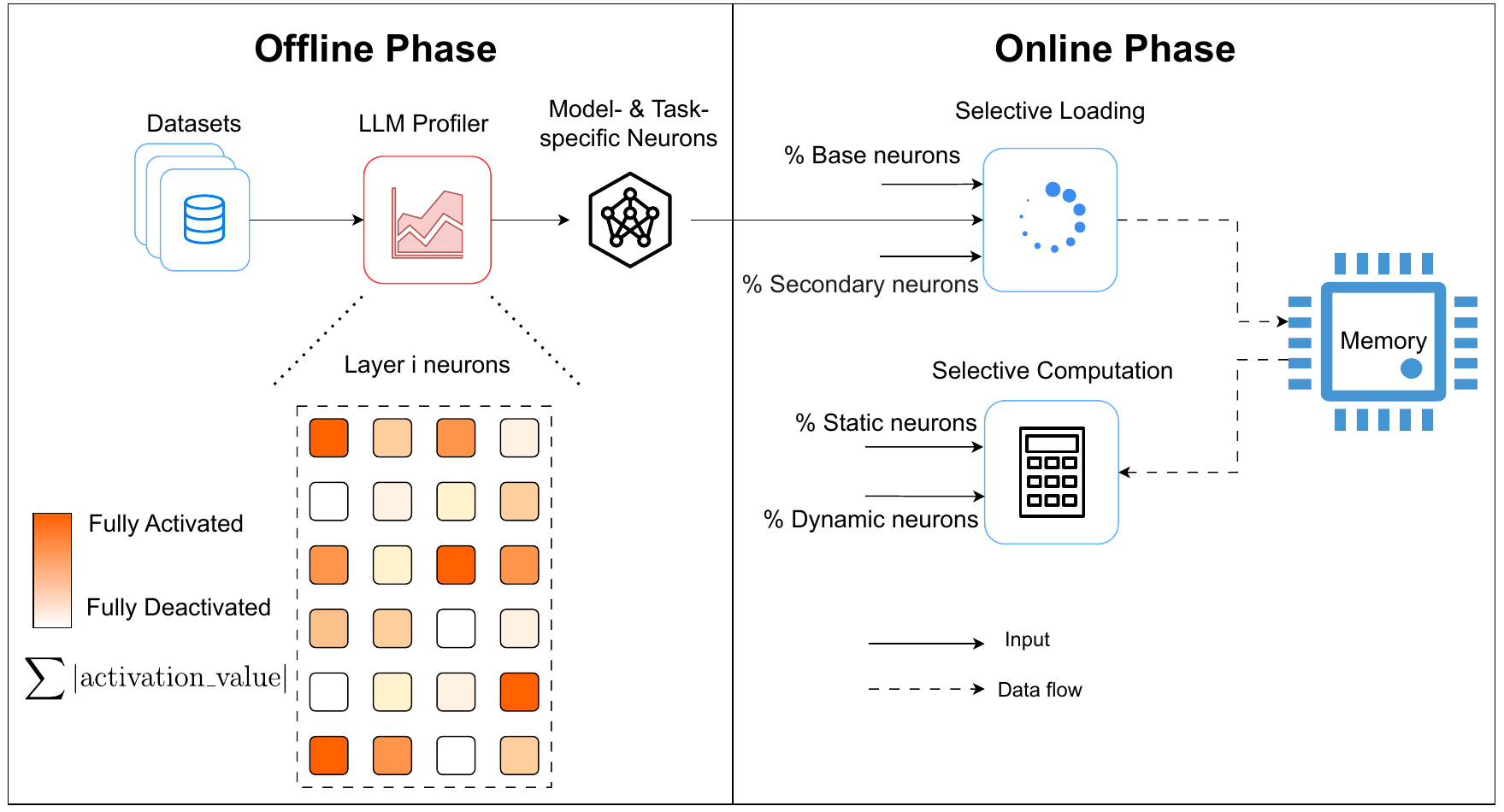}
\caption{SelectInfer has two phases: offline (left) and online (right). In the offline phase, we run a profiler to generate neuron files. In the online phase, selective loading reduces memory usage, while selective computation lowers computational overhead.}
\label{fig:Overview}
\end{figure*}

Despite active research, existing approaches fundamentally fail to optimize the three dimensions  that jointly determine whether an LLM can be practically deployed on an edge device:  \textit{memory footprint}, \textit{inference speed}, and \textit{output accuracy}. Techniques such as quantization reduce memory usage but degrade decoding throughput  and offer only coarse control; pruning methods restore neither memory nor speed without  costly retraining; and neuron-sparsity approaches that operate purely at runtime still require  the full model to reside in memory. On resource-constrained devices, optimizing only one or two of these  dimensions is insufficient: a model that fits in memory but runs too slowly is not deployable,  and a fast model that cannot fit at all is equally useless. What is needed is a framework that  treats all three as first-class concerns and exposes configurable trade-offs between them.

Our key insight is that LLM neurons exhibit a dual structure: a stable core of \textit{model-specific} neurons that activate universally across tasks, and a set of \textit{task-specific} neurons that drive performance for individual inputs. By identifying these two groups before runtime, we can make precise, principled decisions about which neurons to load into memory and which to compute at runtime, treating memory and computation as two independently tunable knobs rather than a single coarse dial. We propose \textbf{SelectInfer}, which identifies active neurons before runtime. Following CoreInfer’s observations~\citep{wang2024coreinfer}, we categorize neurons into two groups: \emph{base neurons}, which activate selectively depending on the input’s task ( e.g., QA, translation, or summarization), and \emph{secondary neurons}, which serve general purposes and activate broadly across inputs. Our approach is summarized in Fig~\ref{fig:Overview}. As shown on the left side, using an offline \textbf{LLM profiler}, we accurately determine these neuron groups. To address memory constraints, in the online phase (right side of Fig~\ref{fig:Overview}), we implement \textbf{selective loading}, a mechanism that loads only the relevant neurons identified by the offline profiler into memory, discarding the remaining ones. Additionally, we introduce \textbf{selective computation}, which selectively computes only the most relevant neurons during runtime, further accelerating inference.

We validate SelectInfer across three lightweight LLMs, Llama3.2-3B~\citep{meta-llama3.2-3b}, Llama3.2-1B~\citep{meta-llama3.2-1b}, and Qwen2.5-3B~\citep{qwen2.5}, on the NVIDIA Jetson Orin Nano~\citep{nvidia-jetson-orin-nano}, a highly memory-constrained device with 8\,GB of shared CPU--GPU memory. Including framework overhead, 3B-scale models demand roughly 8.5--9.5\,GB of memory, but the chip exposes only about $\sim$6.5\,GB of usable memory because the remainder is occupied by the system. As a result, native inference on the chip is not feasible. With \emph{selective loading}, SelectInfer enables both 3B-scale models to run by loading only the most relevant neurons identified during offline profiling. Across QA, translation, and summarization tasks, SelectInfer maintains competitive accuracy while yielding substantial speedups; for example, Llama3.2-3B reaches about 11 tokens/s, achieving 1.53$\times$ improvement over 4-bit quantization, and more than a 13$\times$ over disk-offloading. We observe similar gains for Llama3.2-1B and Qwen2.5-3B, demonstrating that SelectInfer generalizes across model families and offers a practical path to deploy LLMs on resource-constrained edge devices.
% The source code for our approach is publicly available at https://github.com/Huzifa1/SelectInfer.

In summary, our contributions include an offline profiler that identifies model- and task-specific neurons, a novel selective loading scheme that enables deployment of models beyond memory limits on edge devices, and a selective computation mechanism that selectively processes relevant neurons to accelerate inference with negligible accuracy degradation.

\begin{figure*}[t]
\centering
\begin{minipage}[t]{0.30\textwidth}
    \centering
    \includegraphics[width=\linewidth]{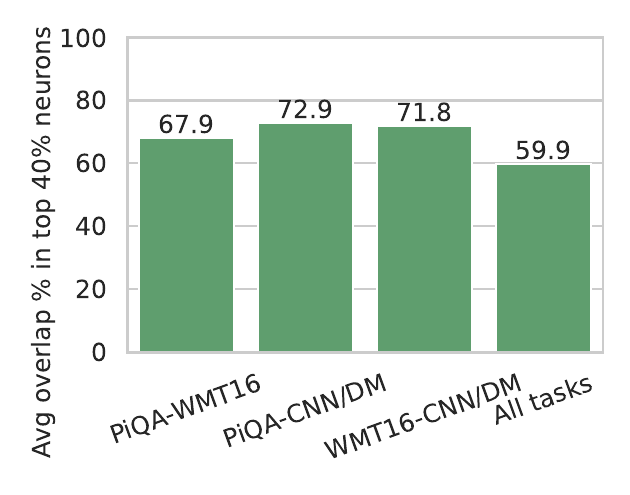}
    \captionof{figure}{High neurons overlap between different tasks, proving that a set of neurons always gets activated, regardless of the input task.}
    \label{fig:model_neurons}
    \vspace{-0.25cm}
\end{minipage}
\hfill
\begin{minipage}[t]{0.30\textwidth}
    \centering
    \includegraphics[width=\linewidth]{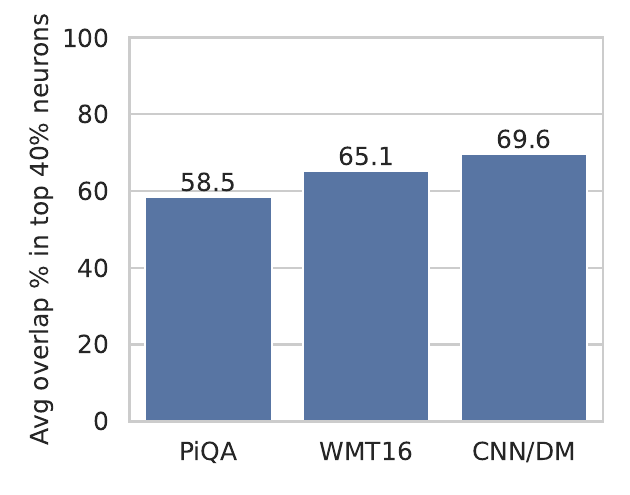}
    \captionof{figure}{High neurons overlap within the same task showing a consistent set of neurons always get activated.}
    \label{fig:task_specific_overlap}
    \vspace{-0.25cm}
\end{minipage}
\hfill
\begin{minipage}[t]{0.30\textwidth}
    \centering
    \includegraphics[width=\linewidth]{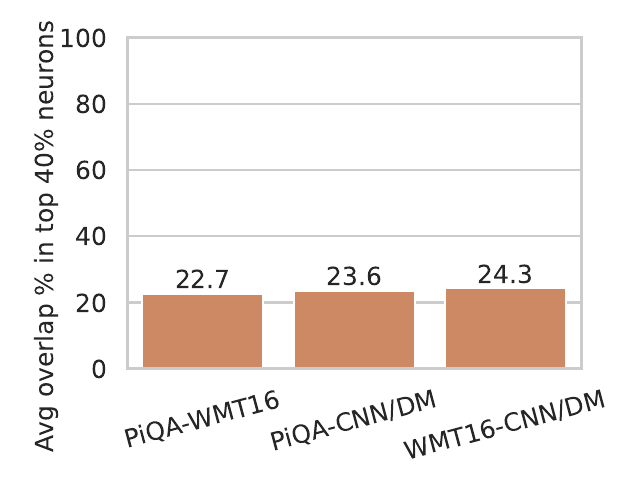}
    \captionof{figure}{Low cross-task neurons overlap. This proves that each task activates its own set of neurons.}
    \label{fig:cross_task_overlap}
    \vspace{-0.25cm}
\end{minipage}
\end{figure*}

\section{Background and Motivation}
\label{sec:key-insight}
This section highlights some background and motivates the insights that shaped the development of our approach.

\subsection{Sparsity in \glspl{llm}}

A key property of \glspl{llm} is their inherent \textit{sparsity}: a small subset of neurons is strongly activated for a given input, largely determining output quality~\citep{liu2023deja, alizadeh2024llm, mirzadeh2023relu}. Thus, the model’s full capacity is rarely used at once; instead, different parts become selectively active depending on the task or input distribution. Prior work has exploited this property at runtime: most notably, CoreInfer~\cite{wang2024coreinfer} observes that activation patterns cluster by semantic domain and identifies active neurons during the prefill phase of each prompt, achieving high throughput by skipping computation over inactive neurons. However, CoreInfer requires the full model to reside in memory and relies entirely on the current prompt for neuron selection. Building on this observation, we further analyze neuron roles across tasks, distinguishing task-specific from model-specific neurons.

\subsection{Model-Specific Neurons}
\label{sec:model-specific-neurons}
We define \textit{model-specific neurons} as neurons that are consistently activated across tasks, regardless of input type. To quantify this, we aggregate prompts from three distinct datasets: PiQA~\citep{piqa} for question answering, CNN/DailyMail~\citep{cnn1,cnn2} for summarization, and WMT16 DE-EN~\citep{wmt16} for translation. Then we measure neuron activations across 100 prompts per task using Llama3.2-3B~\citep{meta-llama3.2-3b} model. We then select the top 40\% of neurons (ranked by absolute activation value) for each layer and examine their overlap across tasks. We average the results across layers. We refer to \S\ref{sec:evalutaion} for more details on the setup. The results are shown in Fig~\ref{fig:model_neurons}. Each bar represents the overlap of the top 40\% neurons for each task pair. We observe a significant cross-task overlap of approximately 68\% to 72\% within the top 40\% of neurons, and approximately 60\% overlap between all three datasets. This provides an evidence that a stable core of neurons is universally engaged across tasks. These neurons represent general-purpose computation in the model and may correspond to linguistic or representational features that are task-agnostic.

\subsection{Task-Specific Neurons}
\label{sec:task-specific-neurons}
In addition to general-purpose neurons, we find strong evidence for \textit{task-specific neurons}. These are neurons that consistently activate within a given task but differ across tasks. To quantify this, for each dataset, we create ten chunks of ten prompts each, identify the top 40\% most active neurons per chunk, and measure the average overlap among all chunks. The results are averaged across all layers. As shown in Fig~\ref{fig:task_specific_overlap}, the overlap ranges from 58.6\% for PiQA to 69.5\% for CNN/DM. This indicates that for each task, a subset of neurons is consistently responsible for processing task-specific information.  
Furthermore, to confirm that these subsets are distinct across tasks, we pick the top 40\% neurons for each task, and calculate the overlap across tasks. As shown in Fig~\ref{fig:cross_task_overlap}, the cross-task overlap is around 24-26\% which is far lower than same-task overlap. This highlights that although each task reuses some task-specific neurons, it also recruits its own specialized subset of neurons.

These results show a dual structure: a set of model-specific neurons that engage universally across tasks, and a set of task-specific neurons that drive performance for individual tasks. This interplay between generality and specialization is a key property of LLM computation in SelectInfer.

\section{\gls{llm} Profiler}
\label{fig:scoring-workflow}
In the offline stage, we run selected datasets through the model to measure neuron activations. In Transformer \gls{ffn} blocks (e.g., \texttt{gate\_proj}, \texttt{up\_proj}, \texttt{down\_proj} in Llama3 and Qwen2.5), each neuron corresponds to a column in the gate/up projections and a row in the down projection. During profiling, we record the absolute activations at the input of the \texttt{down\_proj} layer (i.e., the output of the activation function), which directly correspond to these \gls{ffn} neurons. For each prompt, we record the absolute values of neuron activations, along with the number of input and generated tokens. Since activation functions vary across architectures, the scoring method must sometimes be adapted. For models using SwigLU, we define the following scoring function: Let $\mathcal{T}$ be the set of tokens processed during scoring and let $a_n(t)$ be the activation of neuron $n$ on token $t \in \mathcal{T}$. Then, for each neuron $n \in \{1, \dots, N\}$, we define $s_n$ as the score of the neuron:
\begin{equation}
s_n = \sum_{t \in \mathcal{T}} \left| a_n(t) \right|.
\end{equation}
The scores of all neurons, along with the token counts, are stored in score files. These files can be merged in different configurations by normalizing scores by token count and summing the normalized values. Task-specific neurons can be obtained by merging score files from the same task, while model-level neurons can be obtained by merging across tasks. The final output is a file containing neuron indices ranked in decreasing order of importance across the evaluated datasets.

\section{Selective Loading}

Following the analysis conducted by the \gls{llm} Profiler, the extracted insights are used to optimize the deployment of \glspl{llm} on edge devices by reducing their computational and memory requirements.
We first focus on minimizing the memory footprint, which allows larger models to be deployed on resource-constrained devices where full model loading is otherwise infeasible.

The core idea is to selectively load the model into memory. We apply selective loading to the \gls{ffn}, as this component accounts for the largest proportion of the total model parameters. To reduce the memory usage, we selectively load only a subset of its neurons, focusing on those considered most relevant. This means that, for every block in the \gls{ffn} (e.g., \texttt{gate\_proj}), we select the most important neurons, and load the corresponding slices into the memory, and discard the rest. The remaining components of the model are fully loaded. This selective loading is guided by the neuron relevance profiled by the \gls{llm} Profiler during the offline phase.

\begin{figure*}[t]
\centering
\begin{minipage}[t]{0.45\textwidth}
    \centering
    \includegraphics[width=1\columnwidth]{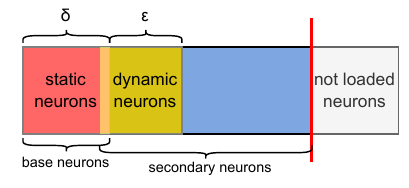}
    \caption{Process of selective computation.}
    \label{fig:partial-computation}
\end{minipage}
\hfill
\begin{minipage}[t]{0.51\textwidth}
    \centering
    \includegraphics[width=1\columnwidth]{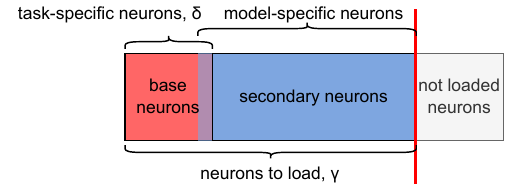}
    \caption{Neurons to load during selective loading.}
    \label{fig:partial-loading-example}
\end{minipage}
\end{figure*}

Our method classifies \gls{ffn} neurons into two groups, \textit{base neurons} and \textit{secondary neurons}. As illustrated in Fig~\ref{fig:partial-loading-example}, we first select the top $\delta$ most relevant \textbf{task-specific} neurons to form the base neurons, and then add the most relevant \textbf{model-specific} neurons as secondary neurons until the overall loading threshold $\gamma$ is reached. We specify the proportion of neurons to be loaded based on memory constraints and application requirements. This selection is applied to each \gls{ffn} layer in the model. When loading the model and reading neurons from disk, we only load the weights corresponding to these selected neurons into memory, leading to substantial memory savings while maintaining competitive accuracy.

\section{Selective Computation}
While selective loading effectively alleviates memory constraints by keeping only a subset of neurons in memory, the computational cost of inference remains significant. To address this, we propose \textit{selective computation}, a runtime mechanism that selectively computes only the most relevant \textbf{loaded} neurons. This strategy reduces computational overhead and accelerates inference without significantly degrading output quality.

As shown in \S\ref{sec:key-insight}, a relatively small fraction of neurons is sufficient to preserve model quality if they are chosen judiciously.
For example, even though $\gamma$ of neurons may be loaded into memory due to selective loading, we can further reduce computation by computing $\epsilon$ of neurons at runtime ($\epsilon$ $<$ $\gamma$). This trade-off between accuracy and speed is user-configurable, enabling flexible deployment depending on application requirements.

During computation, we classify neurons into two categories: static neurons and dynamic neurons. \textbf{Static neurons} are identical to the base neurons selected during selective loading. These task-specific neurons are always computed for every prompt, regardless of the input.
\textbf{Dynamic neurons} are selected at runtime based on the input prompt. Their selection is adaptive and varies per prompt, allowing the computation process to focus on neurons that are most relevant for the current context.

Fig~\ref{fig:partial-computation} illustrates the relationship between static and dynamic neurons in the context of selective computation.
The selection of dynamic neurons occurs during the \textit{prefill phase}, inspired by CoreInfer’s~\citep{wang2024coreinfer} methodology. Fig~\ref{fig:dynamic-neurons} illustrates this process. For each token, we compute activation values for all loaded neurons.
We sort these activations in descending order and select the top $\alpha$ neurons for that token and record the indices of these top $\alpha$ neurons.
This is repeated for all tokens in the prompt, resulting in multiple lists of top neurons.
From these lists, we count the frequency of neuron occurrences and select the top $\beta$ neurons with the highest frequency. These $\beta$ neurons form the dynamic set for this prompt.
In the end, we combine the static neurons with the selected dynamic neurons until the total computation ratio $\phi = \delta + \epsilon$ is reached.
This approach ensures that computation resources concentrates on neurons most likely to contribute to the output for the given input, achieving substantial speedups while retaining accuracy.

\section{Evaluation}
\label{sec:evalutaion}
In this section, we evaluate SelectInfer across the three dimensions central to edge deployment: accuracy (\S\ref{sec:accuracy-results}), memory footprint (\S\ref{sec:memory-results}), and decoding speed (\S\ref{sec:speedup-results}). We first describe the evaluation setup (\S\ref{sec:jetson-board}), the datasets and neuron files used (\S\ref{sec:datasets-selection}), the offline profiler cost (\S\ref{sec:profiler-cost}), and our parameter configuration (\S\ref{sec:param-config}). Throughout, we compare SelectInfer against CoreInfer, 4-bit bitsandbytes quantization, and disk offloading, baselines that each excel on at most one or two of the three target dimensions, to demonstrate that SelectInfer provides a balanced and tunable operating point across all three simultaneously.

\subsection{Setup}
\label{sec:jetson-board}

For our evaluation, we use the NVIDIA Jetson Orin Nano Developer Kit~\citep{nvidia-jetson}, which integrates a 6-core Arm Cortex-A78AE CPU, an NVIDIA Ampere GPU with 1,024 CUDA cores and 32 Tensor Cores, and 8~GB of LPDDR5 shared memory. We benchmark three representative models, Llama3.2-3B, Llama3.2-1B, and Qwen2.5-3B, to compare SelectInfer against two strong baselines: CoreInfer and 4-bit bitsandbytes quantization.

\subsection{Datasets and Neuron-Files Selection}
\label{sec:datasets-selection}
We evaluate three tasks (QA, translation, and summarization) using multiple datasets within each domain. For QA, we use SQuAD~\citep{squad}, TriviaQA~\citep{triviaqa}, MLQA~\citep{mlqa}, and PIQA~\citep{piqa}. For translation, we use WMT16-de-en and WMT16-en-de~\citep{wmt16}, WMT14-fr-en and WMT14-en-fr~\citep{wmt14}. For summarization, we use CNN/DailyMail~\citep{cnn2}, Samsum~\citep{samsum}, and XSum~\citep{xsum}.

We generate task-specific neuron files offline using all datasets within a task, ensuring that the resulting neurons reflects consistent task-specific responses. In contrast, we derive model-specific neurons by applying the same scoring method to a cross-task mixture constructed from one dataset per task, identifying neurons that generalize beyond a single task. Specifically, we use PIQA, WMT14-en-fr, and Xsum datasets to generate model-specific neurons file. We observe that neuron scores stabilize after a few datasets, indicating that this is sufficient for reliable results. Moreover, the choice of datasets for model-specific neuron calculation has little impact on scores, as long as they were drawn from different task groups.

\begin{figure*}[t]
    \centering

    \begin{minipage}[t!]{0.48\textwidth}
        \centering
        \includegraphics[width=1\linewidth]{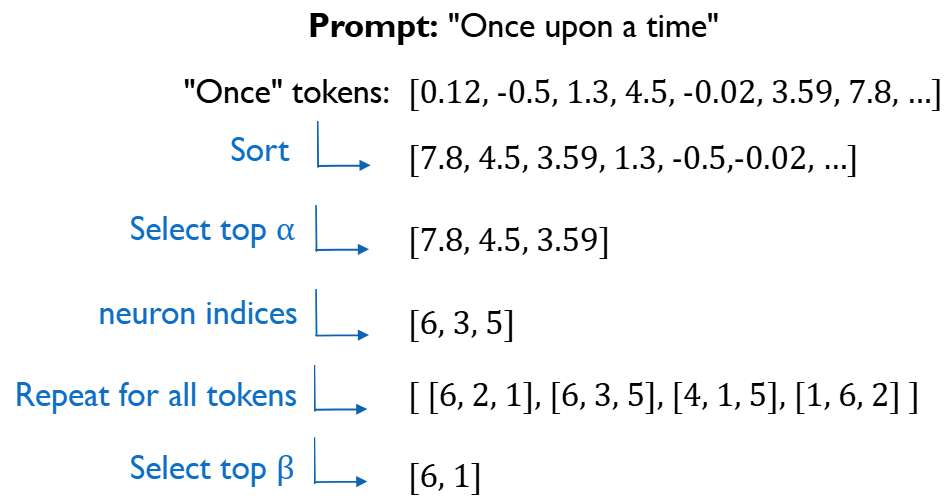}
        \caption{Dynamic neuron selection process.}
        \label{fig:dynamic-neurons}
    \end{minipage}
    \hfill
    \begin{minipage}[t!]{0.48\textwidth}
        \centering
        \captionof{table}{Important parameter values used in our experiments for all three models.}
        \label{tab:parameters}
        \begin{tabular}{lr}
            \toprule
            Parameter & Value \\
            \midrule
            First layer & 5 \\
            Last layer & -2 (Skip last 2) \\
            \% of base / static neurons & 30\% \\
            \% of dynamic neurons & 10\% \\
            \% of computed neurons (sparsity) & 40\% \\
            \% of secondary neurons & 40\% \\
            \% of overall loaded neurons & 70\% \\
            \bottomrule
        \end{tabular}
    \end{minipage}
\end{figure*}

% Bottom: Full-width table
\begin{table*}[t]
    \centering
    \footnotesize
    \caption{Offline profiler cost per model and number of processed tokens.}
    \label{tab:offline-profiler-cost}
    \setlength{\tabcolsep}{4pt}
    \begin{tabular}{lrrrr}
        \toprule
        Model & Tokens processed & Time & Throughput & Peak GPU memory \\
              & (M) & (h) & (tokens/s) & (GB) \\
        \midrule
        Qwen2.5-3B  & 43.9 & 5.51 & 2214 & 16.0 \\
        Llama3.2-3B & 43.7 & 5.01 & 2426 & 21.8 \\
        Llama3.2-1B &  8.8 & 0.56 & 4349 &  7.6 \\
        Llama3.2-1B & 43.8 & 2.69 & 4518 & 10.8 \\
        \bottomrule
    \end{tabular}
\end{table*}

\subsection{LLM Profiler Cost}
\label{sec:profiler-cost}

Table~\ref{tab:offline-profiler-cost} summarizes the offline profiling cost across all models. All results were obtained on an NVIDIA L40S GPU using a batch size of 16. For the 3B-parameter models Qwen2.5-3B and Llama3.2-3B, the profiler processes roughly 44M tokens in 5.0--5.5 hours, achieving throughputs of 2.2k--2.4k tokens/s with peak GPU memory usage between 16~GB and 21.8~GB. For Llama3.2-1B on the same 43.8M-token corpus, profiling completes in 2.69~hours at 4.5k tokens/s with a peak of 10.8~GB GPU memory, while using a smaller 8.8M-token corpus reduces the wall-clock time to 0.56~hours at 4.3k tokens/s and 7.6~GB peak memory. 

\subsection{Parameter Configuration}
\label{sec:param-config}

We find empirically that static neurons should correspond to task-specific neurons, while dynamic neurons represent model-specific behavior. Since task-specific neurons are more directly tied to the target objective, they exhibit higher stability across prompts. For instance, with $\delta = 0.3$, Fig~\ref{fig:task_specific_overlap} shows that roughly $60\%$ of task-specific neurons overlap across prompts within the same task.

SelectInfer exposes several tunable parameters (see Table~\ref{tab:parameters}). Their values depend on model size, device constraints (e.g., memory capacity), expected speedup, and the desired accuracy. In practice, hardware limitations largely determine the feasible percentage of loaded neurons, while the allocation between base and secondary neurons and the fraction of computed neurons must be empirically validated.

For the loading ratio $\gamma$, we observe that $\gamma = 0.7$ is the maximum value at which the 3B-scale models fits on the Jetson Nano chip without running out of memory. To ensure consistent comparison across model sizes, we apply the same value to Llama3.2-1B. The percentage of computed neurons $\phi$ follows prior analyses from CoreInfer, which shows that computing the top $40\%$ of neurons provides a strong accuracy–efficiency trade-off. Our experiments confirmed that this setting works well across tasks and models.

\begin{figure*}[t!]
    \centering
    \begin{minipage}{0.48\linewidth}
        \centering
        \includegraphics[width=0.7\linewidth]{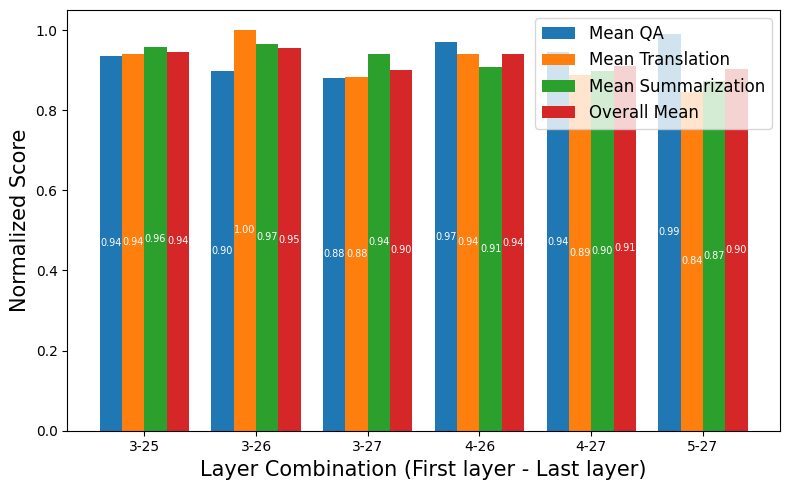}
        \caption{Impact of different layer combinations on cross-task performance. The values are normalized to the maximum value of each task.}
        \label{fig:layer-range-ablation}
        \vspace{-0.2cm}
    \end{minipage}
    \hfill
    \begin{minipage}{0.48\linewidth}
        \centering
        \includegraphics[width=0.7\linewidth]{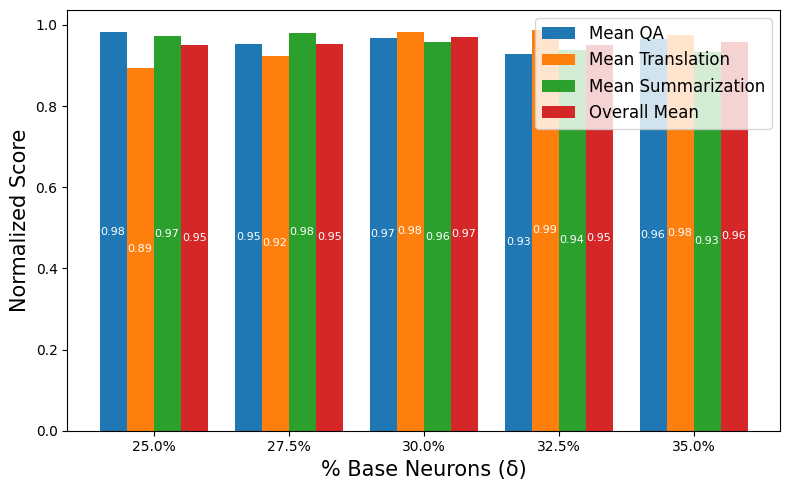}
        \caption{Effect of $\delta$ on performance across different tasks. Values are normalized to the max value of each task.}
        \label{fig:base-percentage-ablation}
        \vspace{-0.2cm}
    \end{minipage}
\end{figure*}

\textbf{Layer-range selection}. We evaluated multiple first/last layer combinations using the Llama3.2-3B model (see Fig~\ref{fig:layer-range-ablation}), which has 28 layers in total, and found that modifying layers 4 through 26 yields the best cross-task performance, considering the number of skipped layers. This aligns with the common observation that early and final layers are more sensitive and have a disproportionate impact on final logits.

\textbf{Division between base and dynamic neurons}. We tested values of $\delta$ from $25\%$ to $35\%$ in small increments (Fig~\ref{fig:base-percentage-ablation}) using the Llama3.2-3B model. The accuracy remains stable across this range, with $\delta = 30\%$ slightly outperforming the alternatives, making it a reasonable and robust choice.

Overall, we choose the parameter values presented in Table~\ref{tab:parameters} based on hardware feasibility, prior validated methodology, and targeted empirical exploration rather than arbitrary selection.

\begin{table*}[t]
\centering
\caption{Benchmark results for SelectInfer and CoreInfer variants across three models and six datasets.}
\label{tab:main-results}
\footnotesize
\resizebox{\textwidth}{!}{
\begin{tabular}{llcccccc}
\toprule
Model & Method & SQuADv2 & MLQA en--en & WMT16 de--en & WMT16 ro--en & XSUM & CNN/DailyMail \\
& (Metric) & (Exact Match) & (Exact Match) & (BLEU) & (BLEU) & (Rouge) & (Rouge) \\
\midrule

\multirow{4}{*}{\textbf{Llama3.2-1B}}
& SelectInfer & \textbf{13.75} & \textbf{0.40} & \textbf{10.50} & \textbf{10.85} & \textbf{0.12} & \textbf{0.15} \\
& CoreInfer & 6.00 & 0.35 & 2.91 & 2.87 & 0.08 & 0.13 \\
& CoreInfer + Selective Loading & 3.85 & 0.28 & 1.68 & 1.39 & 0.05 & 0.11 \\
& CoreInfer + Random Loading & 1.26 & 0.00 & 0.03 & 0.03 & 0.04 & 0.03 \\
\midrule

\multirow{4}{*}{\textbf{Llama3.2-3B}}
& SelectInfer & \textbf{22.55} & 0.52 & \textbf{25.68} & \textbf{25.96} & \textbf{0.12} & \textbf{0.17} \\
& CoreInfer & 17.37 & \textbf{0.53} & 4.97 & 4.98 & 0.00 & 0.15 \\
& CoreInfer + Selective Loading & 11.60 & 0.44 & 1.70 & 1.65 & 0.10 & 0.12 \\
& CoreInfer + Random Loading & 0.26 & 0.01 & 0.04 & 0.03 & 0.00 & 0.05 \\
\midrule

\multirow{4}{*}{\textbf{Qwen2.5-3B}}
& SelectInfer & 24.32 & \textbf{0.53} & \textbf{8.28} & \textbf{4.83} & 0.10 & 0.15 \\
& CoreInfer & 32.58 & 0.53 & 4.32 & 4.09 & \textbf{0.11} & \textbf{0.17} \\
& CoreInfer + Selective Loading & \textbf{35.69} & 0.48 & 1.55 & 0.80 & 0.10 & 0.16 \\
& CoreInfer + Random Loading & 29.65 & 0.34 & 0.44 & 0.16 & 0.07 & 0.13 \\
\bottomrule
\end{tabular}}
\end{table*}

\vspace{-0.2cm}
\subsection{Accuracy}
\label{sec:accuracy-results}

\begin{table*}[t]
\centering
\caption{Benchmark results using Dense and Quantization methods. For dense, we load 70\% of neurons and compute all of them. For quantization, we use 4-bit bitsandbytes~\citep{bitsandbytes}.}
\label{tab:quantized-results}
\footnotesize
\resizebox{\textwidth}{!}{
\begin{tabular}{llcccccc}
\toprule
Model & Method & SQuADv2 & MLQA en--en & WMT16 de--en & WMT16 ro--en & XSUM & CNN/DailyMail \\
& (Metric) & (Exact Match) & (Exact Match) & (BLEU) & (BLEU) & (Rouge) & (Rouge) \\
\midrule

\multirow{2}{*}{\textbf{Llama3.2-1B}}
& 70\% Dense & \textbf{17.87} & \textbf{0.44} & \textbf{27.34} & \textbf{25.95} & 0.14 & 0.16 \\
& 4-bit Quantized & 13.97 & 0.43 & 26.70 & 24.65 & \textbf{0.15} & \textbf{0.17} \\
\midrule

\multirow{2}{*}{\textbf{Llama3.2-3B}}
& 70\% Dense & 26.41 & 0.55 & 34.91 & 33.40 & 0.21 & 0.18 \\
& 4-bit Quantized & \textbf{30.14} & \textbf{0.59} & \textbf{37.64} & \textbf{35.03} & \textbf{0.22} & \textbf{0.19} \\
\midrule

\multirow{2}{*}{\textbf{Qwen2.5-3B}}
& 70\% Dense & 31.06 & 0.58 & \textbf{27.97} & \textbf{20.29} & \textbf{0.14} & \textbf{0.18} \\
& 4-bit Quantized & \textbf{34.32} & \textbf{0.59} & 18.18 & 11.30 & 0.12 & 0.17 \\
\bottomrule
\vspace{-0.5cm}
\end{tabular}}
\end{table*}

Table~\ref{tab:main-results} summarizes the accuracy of our proposed SelectInfer method compared to three baselines: the original CoreInfer~\citep{wang2024coreinfer}, an enhanced version incorporating our selective loading (CoreInfer + Selective Loading), and a control condition where we load neurons at random (CoreInfer + Random Loading). We evaluate these methods on six datasets across three model families: Llama3.2-1B, Llama3.2-3B, and Qwen2.5-3B. The results highlight the complementary benefits of SelectInfer, and demonstrate that its advantages generalize across model families.

\textbf{Superiority of SelectInfer.}
Across models and datasets, SelectInfer mostly achieves the highest accuracy.  
For both Llama models, SelectInfer reaches substantially higher scores than CoreInfer, particularly on translation tasks WMT16 DE--EN and RO--EN. On the 3B model, SelectInfer attains scores around $25$--$26$ BLEU, compared to only $4$--$5$ BLEU for CoreInfer. On the 1B model, SelectInfer reaches $10$--$11$ BLEU, far above CoreInfer’s $2$--$3$ BLEU range. 
For Qwen2.5-3B model, the results further confirm this trend. SelectInfer achieves the strongest performance on both translation tasks ($8.28$ BLEU on DE--EN and $4.83$ BLEU on RO--EN), and remains competitive on MLQA en--en, and both summarization tasks, where it matches or closely trails CoreInfer. The only exception is SQuADv2 dataset, where SelectInfer achieves less accuracy (24.32) than CoreInfer (32.58). Notably, CoreInfer + Selective Loading (35.69) outperforms CoreInfer, eventhough it loads only 70\% of neurons. Another example is in Llama3.2B model in Xsum dataset, where CoreInfer + Selective Loading, achieves 0.1, while CoreInfer achieves 0. We attribute this to the offline profiling stage acting as a beneficial pre-filter: by only loading neurons that are consistently activated across the profiling tasks, some task-irrelevant neurons may be removed, helping CoreInfer focus on a more relevant subset.

\textbf{Effectiveness of selective loading.}
Comparing CoreInfer + Selective Loading against CoreInfer + Random Loading isolates the benefit of profiling-based neuron selection.  
Random loading severely damages performance in every setting, often collapsing accuracy to near-zero values across all six tasks. In contrast, selective loading maintains a meaningful proportion of CoreInfer’s performance, even though only $\gamma = 70\%$ of neurons are available. This stability holds across all evaluated models. On Qwen2.5-3B, for example, CoreInfer + Selective Loading achieves BLEU scores of $1.55$ and $0.80$ on the translation tasks, whereas random loading yields only $0.44$ and $0.16$. Although selective loading naturally reduces performance relative to full-neuron CoreInfer, it clearly preserves more accuracy than random selection. This demonstrates that profiling is essential for identifying consistently active and task-relevant neurons.

\textbf{Contribution of selective computation.}
To understand the effect of selective computation, we compare SelectInfer with CoreInfer + Selective Loading, which shares the same set of loaded neurons but differs during computation. Across all models and nearly all datasets, SelectInfer consistently outperforms its selective-loading counterpart. The gaps are particularly large on the translation datasets: on Llama3.2-3B, SelectInfer achieves $25.68$ BLEU on WMT16 DE--EN compared to only $1.70$ for CoreInfer + Selective Loading, and similar patterns appear on Llama3.2-1B and Qwen2.5-3B. These results demonstrate that profiling-guided selective computation is crucial for identifying and prioritizing the most informative neurons during inference.

\textbf{Comparison with dense and quantized baselines.}
Table~\ref{tab:quantized-results} extends our analysis by evaluating an additional baseline, \textbf{4-bit quantized} variant using bitsandbytes~\citep{bitsandbytes} quantization, which is commonly used to reduce inference cost and memory footprint. To make it a fair comparison, we compare it against 70\% dense configuration, where only 70\% of neurons are loaded using our selective loading, but all of them are fully computed.
Across all three models, the 70\% dense baseline proves to be a strong and reliable reference point, often outperforming or matching the 4-bit quantized version, especially on translation datasets. 
For Llama3.2-1B, the dense baseline dominates on WMT16 DE--EN and RO--EN, reaching $27.34$ and $25.95$ BLEU, whereas the quantized version achieves slightly lower values ($26.70$ and $24.65$).  
This pattern persists in the Qwen2.5-3B model, where dense loading achieves substantially stronger translation performance ($27.97$ BLEU on DE--EN and $20.29$ on RO--EN), while quantization drops to $18.18$ and $11.30$ BLEU.  
On the other hand, the quantized version performs competitively on QA tasks (e.g., MLQA and SQuADv2), and occasionally surpasses the dense version on Llama3.2-3B and Qwen2.5-3B. However, even in these settings, the differences are generally small, and the dense baseline remains competitive and stable across all tasks.

\textbf{Summary.}
These results validate the dual contribution of our mechanisms: selective loading improves memory efficiency without degrading performance, while selective computation boosts performance by focusing on the most relevant neurons. Additionally memory footprint and speedup can be controlled independently. Available memory can be fully utilized while speedup can be aligned to available computational resources and application requirements.

\subsection{Memory Footprint}
\label{sec:memory-results}

A key motivation behind SelectInfer is reducing the memory required to load and execute \gls{llm} on resource-constrained edge devices. Unlike full-parameter loading approaches, SelectInfer stores only a subset of neurons in memory. The theoretical reduction in memory usage can be expressed as
\begin{equation}
\label{eq:memory}
    \Delta M = N_{\text{ffn}} \times (1 - r_{load}) \times r_{layer} \times s,
\end{equation}
where $N_{\text{ffn}}$ denotes the number of FFN parameters, $r_{load}$ is the fraction of neurons preloaded into memory, $r_{layer}$ is the fraction of layers where SelectInfer is applied, and $s$ is the storage size per parameter. For example, in Llama3.2-3B, where $N_{\text{ffn}} \approx 2.1$B, choosing $r_{load}=0.7$, $r_{layer}=0.75$, and $s=2$ bytes (\texttt{float16}) yields $0.945\ \text{GB}$, aligning with the value observed during deployment.

\begin{table*}[t]
\centering
\caption{End-to-end peak memory usage (in GB) for different inference approaches across three models.}
\label{tab:memory-usage}
\begin{tabular}{lcccc}
\toprule
\textbf{Model} & \textbf{Original} & \textbf{SelectInfer} & \textbf{CoreInfer} & \textbf{Quantized (4-bit)} \\
\midrule
Llama3.2-3B   & 6.88 & 5.97 & 7.44 & 3.03 \\
Llama3.2-1B   & 2.86 & 2.60 & 2.86 & 1.52 \\
Qwen2.5-3B    & 6.48 & 5.37 & 7.52 & 2.67 \\
\bottomrule
\end{tabular}
\end{table*}

To validate these analytical expectations, we report the end-to-end memory usage of SelectInfer against three baselines: the original model (full-parameter loading), CoreInfer, and 4-bit bitsandbytes quantization. Table~\ref{tab:memory-usage} summarizes the results.
As expected, the original model incurs full memory footprint, since all neurons are loaded in memory during inference. CoreInfer consistently exceeds the reference memory usage due to additional buffers and intermediate structures needed to support its core neurons computation. In contrast, SelectInfer achieves a lower memory footprint across all evaluated models as it loads only a fraction of neurons into memory (70\% in our experiments).
The reduction is most noticeable in the 3B-scale models. For Llama3.2-3B, SelectInfer reduces memory from 6.88 GB to 5.97 GB, corresponding to a savings of 13.2\%, consistent with the analytical estimate in Eq~\ref{eq:memory}. A similar trend appears in Qwen2.5-3B, where SelectInfer achieves a reduction of over 1.1 GB relative to the original model.
For the 1B model, the improvement is smaller: from 2.86 GB to 2.6 GB. This is because we apply SelectInfer to fewer layers in Llama3.2-1B (9 of 16 layers), reducing the relative impact of selective loading compared to larger models where a higher proportion of layers are covered.
Finally, 4-bit quantization unsurprisingly yields the lowest memory footprint overall, as it compresses all weights, whereas SelectInfer selectively loads fewer neurons but retains full precision for the loaded parameters. However, quantization trades off memory reduction for lower decoding throughput, as we will discuss in the next subsection.

\subsection{Speedup}
\label{sec:speedup-results}

In this section, we evaluate the impact of SelectInfer on decoding throughput (tokens/s) and task accuracy using Llama3.2-3B and Llama3.2-1B models.
The decoding speed corresponding to varying proportions of loaded neurons is illustrated in Fig~\ref{fig:partial_loading}.
Reducing the fraction of loaded neurons leads to an increase in prefilling and decoding stage, since fewer weights are fetched from memory and computed.
In our example, loading only $\gamma = 40\%$ of the neurons improves throughput from $8.65$ token/s to $9.75$ tokens/s compared to 80\% loading. However, this comes at the cost of accuracy: on SQuADv2, accuracy improves from 14.87 at 40\% loaded neurons to 24.70 at 80\%.

\begin{figure*}[t!]
    \centering
    \begin{minipage}{0.48\linewidth}
        \centering
        \includegraphics[width=1\columnwidth]{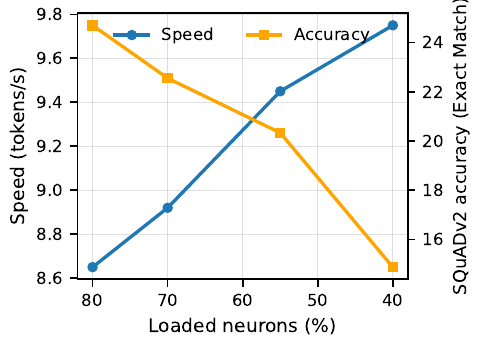}
    \caption{Effect of selective loading on decoding speed and accuracy using Llama3.2-3B.}
        \vspace{-0.2cm}
    \end{minipage}
    \hfill
    \begin{minipage}{0.48\linewidth}
        \centering
        \includegraphics[width=1\columnwidth]{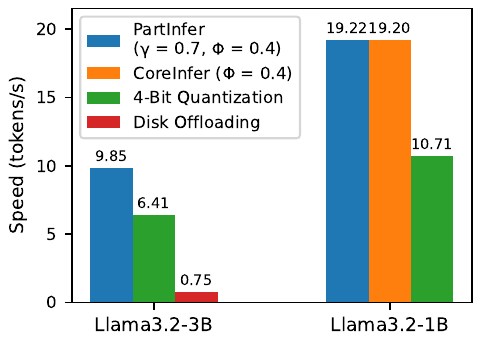}
    \caption{Throughput comparison between baselines.}
    \label{fig:comparison}
    \end{minipage}
\end{figure*}

\textbf{Comparison with other baselines}. 
We evaluate the decoding speed of SelectInfer, against three baselines on the NVIDIA Jetson Orin Nano platform: CoreInfer, 4-bit bitsandbytes~\citep{bitsandbytes} quantization, and disk offloading. Fig~\ref{fig:comparison} summarizes these results. For the Llama3.2-3B model, SelectInfer attains 9.85 tokens/s, outperforming 4-bit quantization (6.41 tokens/s) by roughly 1.5$\times$ and exceeding disk offloading (0.75 tokens/s) by more than 13$\times$, the latter being heavily bottlenecked by frequent parameter transfers between disk and GPU memory. CoreInfer is excluded from the 3B comparison because it cannot fit within the memory constraints of the device. The substantial speedup of SelectInfer arises from selectively computing only 40\% of neurons at runtime, in contrast to quantized inference, which evaluates all neurons. For the Llama3.2-1B model, SelectInfer and CoreInfer achieve comparable performance ($\sim$19.2 tokens/s), as both similarly restrict computation to 40\% of neurons. Meanwhile, the quantized model reaches only 10.71 tokens/s, approximately half the speed of the previous approaches.

\subsection{Summary}
Our evaluation shows that SelectInfer provides a balanced and tunable trade-off across accuracy, memory usage, and decoding speed, metrics that existing approaches struggle to optimize simultaneously. Although 4-bit quantization achieves the smallest memory footprint (about half of SelectInfer’s; Table~\ref{tab:memory-usage}), SelectInfer reaches comparable accuracy while delivering \textbf{1.5--2$\times$} higher decoding throughput (Fig~\ref{fig:comparison}), and it avoids the hardware limitations of low-precision kernels, which are not consistently supported across GPUs~\citep{nvidia2025tensorrt4bit, doubleword2024quantization}. Moreover, quantization offers only coarse control (typically 4-bit vs.\ 8-bit), making it difficult to fully utilize the available memory or computation of edge devices. In contrast, SelectInfer exposes fine-grained knobs that allow users to adjust the percentage of loaded and computed neurons, enabling precise memory–speed–accuracy trade-offs and allowing the model to scale up or down to match device capabilities. Compared to CoreInfer, with fast decoding speed but less accurate and no memory savings, SelectInfer provides a more robust and adaptable framework.

\section{Related Work}

\paragraph{Quantization and compression.}
Quantization reduces model size by representing weights at lower precision, and has been widely adopted to shrink LLMs for deployment~\cite{dettmers2023qlora, lin2024awq}. While effective at reducing memory footprint, quantization degrades output quality and, counterintuitively, can \textit{reduce} decoding throughput on hardware that lacks native support for low-precision kernels~\cite{kim20213,nvidia2025tensorrt4bit}. Furthermore, it offers only coarse granularity (typically 4-bit or 8-bit), making it difficult to precisely match a model's resource consumption to available device capacity. Model compression via weight sharing or knowledge distillation~\cite{choudhary2020comprehensive} similarly requires expensive retraining and provides limited runtime flexibility.

\paragraph{Pruning.}
Structured and unstructured pruning methods~\cite{bansal2022rethinking, ma2023llm} reduce the number of active parameters by removing less important weights or neurons. While they can reduce both memory and computation, they typically require fine-tuning to recover lost  accuracy, which is costly and task-specific. Unlike SelectInfer, pruning permanently removes weights from the model and cannot adapt to varying tasks or memory budgets at deployment time.

\paragraph{Offloading.}
Offloading approaches such as DeepSpeed~\cite{aminabadi2022deepspeed} and Medusa~\cite{cai2023medusa} move portions of the model to CPU or disk, alleviating GPU memory pressure. However, they are fundamentally limited by PCIe bandwidth, require powerful host CPUs, and are incompatible with unified CPU--GPU memory architectures common in edge SoCs such as NVIDIA Jetson~\cite{nvidia-jetson} and Qualcomm Snapdragon~\cite{dagli2024shared}.

\paragraph{Neuron sparsity at runtime.}
A line of work exploits the observation that only a small fraction of neurons strongly activate for any given input~\cite{mirzadeh2023relu, liu2023deja, alizadeh2024llm}. PowerInfer~\cite{song2024powerinfer} and PowerInfer-2~\cite{powerinfer2} train lightweight MLP predictors at each layer to identify active neurons dynamically, but these predictors themselves add approximately 15--20\% memory overhead and impose additional computational cost. LLM-in-a-Flash~\cite{alizadeh2024llm} enables inference from flash storage by fetching only active neuron weights on demand, reducing memory pressure but remaining bottlenecked by storage I/O bandwidth.

\paragraph{Semantics-aware sparse activation.}
CoreInfer~\cite{wang2024coreinfer} is the most closely related prior work. It observes that neuron activation patterns cluster by semantic domain and exploits this by identifying active neurons during the prefill phase of each prompt. This yields high decoding throughput by skipping computation over inactive neurons at runtime. However, CoreInfer requires the \textit{full} model to be loaded into memory and it does not leverage any offline model-level knowledge, relying entirely on the current prompt's prefill to select neurons. As a result, CoreInfer cannot run 3B-scale models on devices with fewer than $\sim$7~GB of usable memory (e.g., NVIDIA Jetson Orin Nano), and its neuron selection is noisier than SelectInfer's offline-profiled approach.

\paragraph{Positioning of SelectInfer.}
SelectInfer differs from all of the above by jointly addressing memory, speed, and accuracy through two complementary mechanisms: \textbf{selective loading} and \textbf{selective computation}, both guided by an \textbf{offline profiler} that identifies model-specific and task-specific neurons. Unlike quantization, it retains full precision for loaded weights and offers fine-grained, continuously tunable memory--speed--accuracy knobs. Unlike pruning, it requires no retraining. Unlike offloading, it is compatible with unified memory architectures. Unlike runtime-only sparsity methods, it reduces memory footprint, enabling models that would otherwise not fit on a device to run successfully.

\section{Conclusion}
This paper presents SelectInfer, a neuron-level optimization framework for deploying \gls{llm} on edge devices.
It provides fine-grained control over memory footprint and computational requirements, while maintaining the accuracy of the generated outputs.
We define model-specific and task-specific neurons highlighting their distinct roles.
We exploit this insight to introduce selective loading to reduce the memory footprint and selective computation to lower computational demands.
These optimizations are realized by model knowledge extracted during an offline phase.
Evaluations demonstrate effectiveness in enabling \gls{llm} inference on resource-constrained devices. Experimental results show the effectiveness of SelectInfer on memory footprint and speedup while maintaining high accuracy.

Future work will extend the current approach beyond CPU and GPU execution by incorporating support for neural processing units, such as those available on Jetson devices. In addition, further research will address the current limitation of manually identifying task-specific neurons offline. This could be addressed by integrating lightweight classifiers that allow dynamic neuron selection during inference based on the task type. Finally, to broaden the applicability of the approach, future efforts will focus on generalizing the method to a wider range of models and tasks.

\newpage

\bibliographystyle{unsrtnat}
\bibliography{refs}

\newpage

\end{document}